\documentclass[reprint,amsmath,amssymb,aps,prb,superscriptaddress,longbibliography]{revtex4-1}

\pdfoutput=1

\usepackage{amsmath,amssymb,mathtools,amsfonts}
\usepackage{graphicx, dcolumn, float, placeins}
\usepackage{siunitx}
\usepackage[version=4]{mhchem}
\usepackage{url}
\usepackage{mciteplus}
\usepackage{comment}
\usepackage{xr}
\usepackage{pdfpages}
\usepackage[utf8]{inputenc}
\DeclareUnicodeCharacter{22EF}{\ldots}
\usepackage{lineno}
\usepackage{xcolor}

\usepackage{hyperref}
\hypersetup{breaklinks,colorlinks=true,linkcolor=blue,citecolor=blue,urlcolor=blue,filecolor=blue}

\usepackage{cleveref}
\makeatletter
\AtBeginDocument{\let\LS@rot\@undefined}
\makeatother

\newcommand\mat\mathbf

\DeclareUnicodeCharacter{22EF}{\ldots}

\newcommand{\byte}
{\affiliation{ByteDance Seed}}
\newcommand{\pku}
{\affiliation{Peking University}}
\newcommand{\thu}
{\affiliation{Tsinghua University}}

\begin{document}
\author{Bangchen Yin}
\byte
\thu

\author{Jian Ouyang}
\byte
\thu

\author{Zhen Fan}
\byte

\author{Kailai Lin}
\byte

\author{Hanshi Hu}
\thu

\author{Dingshun Lv}
\byte

\author{Weiluo Ren}
\byte

\author{Hai Xiao}
\email{haixiao@tsinghua.edu.cn}
\thu

\author{Ji Chen}
\email{ji.chen@pku.edu.cn}
\pku

\author{Changsu Cao}
\email{caochangsu@bytedance.com}
\byte

%
%
\title{Hessian-informed machine learning interatomic potential \\ towards bridging theory and experiments}

\begin{abstract}

Local curvature of potential energy surfaces is critical for predicting certain experimental observables of molecules and materials from first principles, yet it remains far beyond reach for complex systems. 
In this work, we introduce a Hessian-informed Machine Learning Interatomic Potential (Hi-MLIP) that captures such curvature reliably, thereby enabling accurate analysis of associated thermodynamic and kinetic phenomena.
To make Hessian supervision practically viable, we develop a highly efficient training protocol, termed Hessian INformed Training~(HINT), achieving two to four orders of magnitude reduction for the requirement of expensive Hessian labels.
HINT integrates critical techniques, including Hessian pre-training, configuration sampling, curriculum learning and stochastic projection Hessian loss.
Enabled by HINT, Hi-MLIP significantly improves transition-state search and brings Gibbs free-energy predictions close to chemical accuracy especially in data-scarce regimes.
Our framework also enables accurate treatment of strongly anharmonic hydrides, reproducing phonon renormalization and superconducting critical temperatures in close agreement with experiment while bypassing the computational bottleneck of anharmonic calculations.
These results establish a practical route to enhancing curvature awareness of machine learning interatomic potentials, bridging simulation and experimental observables across a wide range of systems.

\end{abstract}

\maketitle
\newpage

\section{Introduction}
Understanding the behavior of chemical systems under realistic conditions from first principles calculations remains a longstanding challenge, especially for systems governed by thermal fluctuations and vibrational dynamics, ranging from catalytic activity to phase stability ~\cite{textbook1, houkSmallmolecule2008, baroneAnharmoinc2004,truhlarVariational1980}. 
Key observables, including vibrational spectra, thermal/electron conductivities, and heat capacities, depend on the local curvature of the potential energy surface (PES) ~\cite{wilsonMolecular1955,togoFirst2015} and are therefore costly to compute.
In practice, the principal bottleneck is the evaluation of the Hessian matrix.
For instance, in standard Density Functional Theory~(DFT), this step typically scales as $O(N^5)$, compared with $O(N^3)$ for total energy calculations ~\cite{gerrattForce1968}. 

Machine learning interatomic potentials~(MLIPs) have recently revolutionized computational chemistry by emulating quantum-mechanical accuracy at a fraction of the cost~\cite{behlerGeneralized2007,bartokGaussian2010,DeepMDkit,schnet, Batatia2022mace,Neuip,nep89,zhangPhase2021,lu862021}.
Yet, a critical limitation remains: standard MLIPs often struggle to capture the delicate curvature of the PES~\cite{wanderAccessingNumericalEnergy2025}.
While these models may predict forces accurately enough for short trajectories, they may lack the second-order accuracy essential for vibrational properties, transition state analysis, and numerical stability for long-timescale dynamics simulations. 
Closing this gap may require moving beyond standard energy~($E$) and force~($F$) training to explicitly include higher-order derivative information.

However, directly training MLIPs on Hessian labels poses a fundamental computational challenge due to the dimensionality of the Hessian matrix ($3N\times 3N$, $N$ being the number of atoms), which is far greater than that of energy scalars or force vectors.
This dimensionality causes severe bottlenecks on data generation process, making high-fidelity Hessian labels via first-principles methods orders of magnitude more expensive to obtain than energy and forces.
Consequently, datasets containing Hessian labels remain limited, including QM9-Hessian~\cite{QM9_Hess}, OpenREACT~\cite{Hessian_MD} and HORM~\cite{HORM} for molecular systems, alongside  MDR~\cite{MDR}, PBE-MDR~\cite{PBE-MDR} for periodic materials.
Beyond the data scarcity, the training complexity itself is also prohibitive. 
Supervising a model directly on full $3N \times 3N$ matrices can be upwards of 25 times more computationally demanding than standard energy-force training~\cite{Hessian_MD,rodriguezProjected2026}. 
Existing strategies to mitigate these costs typically involve dimensionality reduction, such as selecting specific columns of the Hessian matrix in molecular~\cite{HORM} systems or periodic material systems~\cite{PFTphonon}, or employing decoupled readout heads to predict second-order information as an independent task~\cite{HIP}.

In this work, we address these challenges by introducing a Hessian-informed machine learning interatomic potential~(Hi-MLIP) that achieves Hessian-level accuracy while remaining  computationally tractable.
This is enabled by a practical Hessian-INformed Training~(HINT) protocol, which integrates (i) Hessian pre-training, (ii) configuration sampling, (iii) curriculum learning and (iv) stochastic projected Hessian loss. 
This approach achieves model convergence using two to four orders of magnitude fewer high-fidelity Hessian labels than standard training methods. 
Furthermore, we showcase the practical power of Hi-MLIP on real downstream tasks of transition state search and Gibbs free energy calculation in molecules and prediction of the phonon anharmonicity in materials.
Overall, this work establishes a computationally practical framework for incorporating Hessian supervision, providing a step toward closer alignment between atomic simulations and experimental measurements.

\section{Results}
\subsection*{Framework}
\begin{figure*}[ht]
 \centering
 \includegraphics[width=0.85\textwidth]{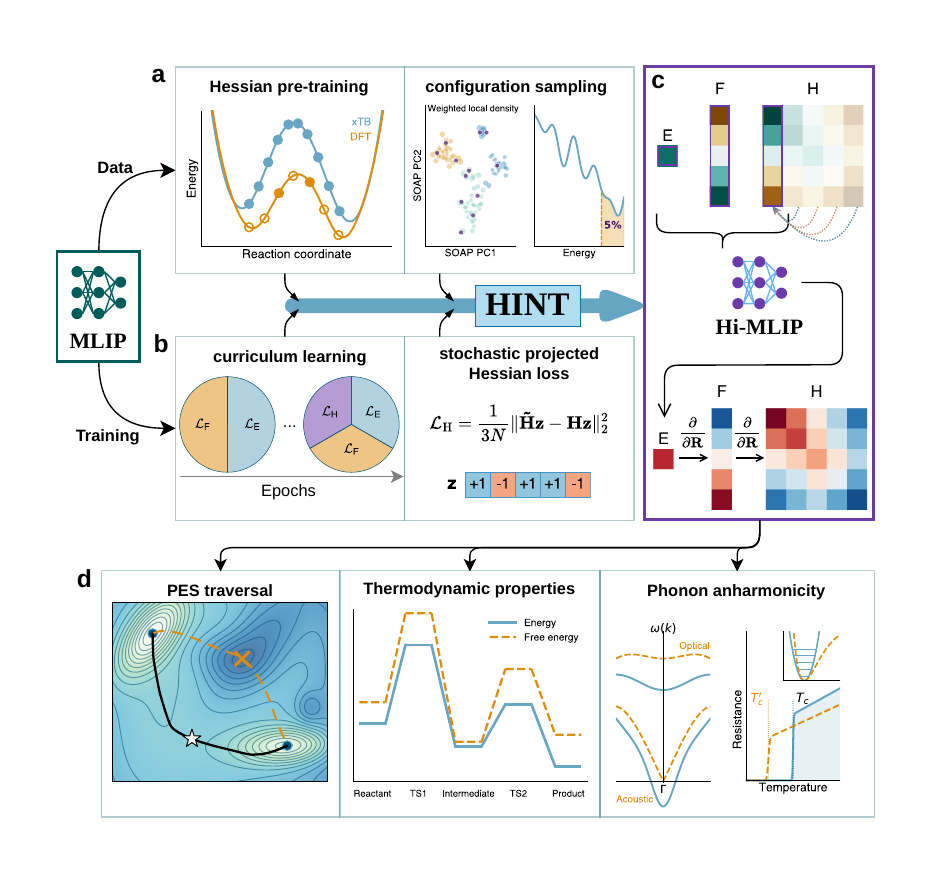}
\caption{\textbf{Framework of the Hessian-informed training (HINT) protocol enables accurate prediction of thermodynamic and kinetic properties.} \textbf{(a)} Efficient data preparation. Hessian pre-training leverages large-scale low-fidelity~(xTB) Hessian data to faithfully learn the global PES curvature. High-fidelity~(DFT) configurations are then sampled via the weighted-local density sampling in SOAP descriptor space or energy-based ranking, where retaining only the top 5\% of configurations by Hessian label achieves comparable performance to the full dataset. \textbf{(b)} Efficient training process. A curriculum learning strategy dynamically reweights the Energy/Force/Hessian labels over training epochs. The stochastic projected Hessian loss, based on the Hutchinson estimation with random $\pm 1$ vectors, achieves $\mathcal{O}(N)$ training efficiency. \textbf{(c)} Hi-MLIP architecture. The HINT protocol augments a base MLIP with Hessian supervision and yields a Hi-MLIP.  Force and Hessian predictions from the Hi-MLIP are obtained by successive analytical differentiation of the potential energy, ensuring physical consistency and improved accuracy over the base MLIP model. \textbf{(d)} Applications enabled by Hi-MLIP, including PES traversal for transition state searches, free energy and other thermodynamic property prediction, and phonon anharmonicity calculation in hydrides under pressure.}
\label{fig:workflow}
\end{figure*}

In this section, we present Hessian-INformed Training~(HINT) protocol, a data- and training-efficient framework integrating Hessian information into MLIP training process.
Our approach systematically minimizes the reliance on expensive reference data while accelerating model convergence 
by jointly improving data efficiency and trianing efficiency.
The four critical components of this protocol are illustrated in Fig.~\ref{fig:workflow} and discussed as follows.
Further details are comprehensively discussed in Sec.~\ref{sec:methods}. 

Our workflow begins with a Hessian pre-training stage~\cite{dpa2,sevenet,crossfunction} designed to improve data efficiency for first-principles Hessian labels. 
Since large-scale, low-fidelity Hessian datasets are relatively easy to construct, they provide an ideal starting point.
We pre-train our MLIP model on such datasets to establish a robust foundational understanding of the global PES curvature. 
For that purpose, we construct two datasets with Hessian data on semi-empirical GFN2-xTB level~\cite{xtb,gfn2-xtb}, namely the T1x-xTB-Hess and HORM-xTB-Hess datasets.

After establishing a low-fidelity representation of PES curvature, we implement a data sampling stage to identify the most informative configurations for the subsequent high-fidelity fine-tuning.
As a general strategy, we apply weighted-local density sampling within the SOAP descriptor space~\cite{SOAP}.
In particular, for transition state searches, we find that energy-based ranking method to be highly effective.
 
As for the high-fidelity fine-tuning stage, we adopt a heterogeneous labeling approach. 
The fine-tuning dataset consists of a large amount of configurations with energy and force~(E/F) labels, augmented by a small fraction of configurations featuring full Hessian~(E/F/H) labels. 
To manage the inherent imbalance of this heterogeneous dataset, we implement a curriculum learning strategy governed by a dynamic weight schedule.
By progressively growing the weight of the scarce Hessian labels relative to the abundant E/F data during training, this schedule effectively prevents overfitting and ensures stable convergence.

Beyond dataset composition, the direct supervision of $3N \times 3N$ Hessian matrices imposes a heavy computational burden during backpropagation.

Instead of explicitly constructing the full Hessian matrix, we utilize a stochastic projection loss.
By leveraging the Hessian-vector product (HvP) algorithm in automatic differentiation frameworks, we can efficiently compute the projection of the Hessian along a random vector $\mathbf{z}$. 
Conceptually, there are two distinct strategies for constructing this projection vector $\mathbf{z}$. 
The conventional method employs coordinate basis vectors, which is effectively equivalent to randomly sampling individual columns of the Hessian, an approach also presented in previous works~\cite{HORM, PFTphonon,rodriguezProjected2026}.
Alternatively, a more advanced strategy is to sample $\mathbf{z}$ from a Rademacher distribution (where elements are $\pm 1$ with equal probability), consistent with Hutchinson estimation~\cite{hutchinsonStochastic1989}. 
This allows the model to capture error signals across all dimensions simultaneously rather than sparsely.
This stochastic formulation captures the global PES curvature more effectively than single-column projections as we will show in the following subsection.

Together, these components constitute a general HINT protocol that enables practical Hessian supervision in MLIP training, as summarized in Fig.~\ref{fig:workflow}.
In the following subsections, we demonstrate the effectiveness of resulting Hi-MLIPs on thermo-kinetic and anharmonic phenomena.

\subsection*{Thermo-kinetic Properties in Molecules}\label{sec:mol_react}
\begin{figure*}[ht]
 \centering
 \includegraphics[width=0.95\textwidth]{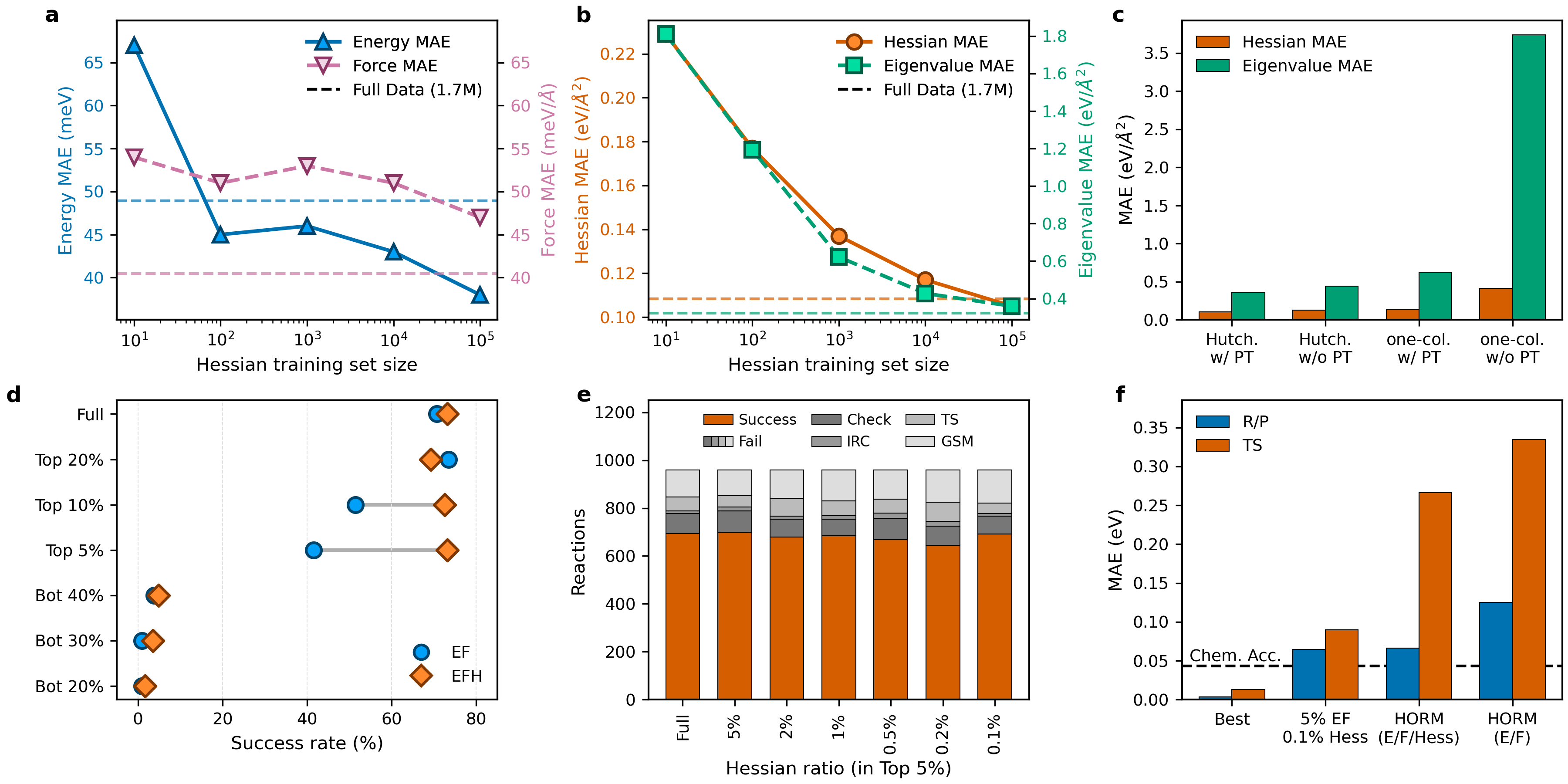}
\caption{\textbf{Data efficiency and predictive accuracy of the Hessian-informed training protocol on molecular reactions.} \textbf{(a, b)} Label efficiency of the protocol, showing MAEs for (a) energy/forces and (b) Hessian/eigenvalues with increasing Hessian training samples. Horizontal dashed lines denote the full-dataset baseline~($\sim$1.7M configurations).\textbf{(c)} Ablation study comparing loss functions and pre-training strategies (PT) using ${10^5}$ Hessian samples. \textbf{(d)} Success rates of transition state (TS) searches trained on varying energy-based subsets (high-energy Top 5--20\% vs. low-energy Bot 20--40\%), comparing models trained with (EFH) and without (EF) Hessian labels. \textbf{(e)} TS search success rate versus Hessian data ratio within the Top 5\% high-energy subset. \textbf{(f)} Generalization performance on free energy prediction. The average MAE for 124 randomly selected reactant/product (R/P) and transition state (TS) structures are compared across different models. The best model achieves the highest accuracy. The 5\% EF + 0.1\% Hess~(0.0005\% Hess of full dataset) model demonstrates remarkable data efficiency, outperforming the HORM baselines and approaching chemical accuracy~(black dashed line, 0.043 eV).} 
\label{fig:mol_react}
\end{figure*}
Identifying transition states and reaction pathways on PES is essential for predicting the rates and thermodynamic properties of chemical reactions. To evaluate our HINT protocol in this scenario, we utilized the HORM dataset~\cite{HORM}, which has ${\sim}$1.7 million explicit DFT Hessian labels. This dataset provides an ideal platform for systematically benchmarking the accuracy and scaling behavior of our protocol.

As illustrated in Fig.~\ref{fig:mol_react}a,b, we first investigate the data efficiency of our protocol using AlphaNet~\cite{alphanet} as the base MLIP architecture. By pre-training on HORM-xTB-Hess, fixing a fine-tuning subset of 100,000 configurations and varying the number of provided Hessian labels from 10 to 100,000, we evaluate the model performance versus the amount of high-fidelity DFT Hessian labels. For comparison, we establish a full-dataset baseline~(dashed lines) using projection Hessian loss in Hutchinson way but without low-fidelity Hessian pre-training. While the mean absolute errors~(MAEs) for energy and forces remain highly stable across the sampled range, the Hessian and eigenvalue MAEs decrease significantly as more DFT Hessian labels are introduced. Notably, our model shows convergence and achieves similar accuracy with the full-dataset baseline~($\sim$1.7M) using only 10,000 DFT Hessian labels, representing a $\sim$170-fold reduction in high-fidelity Hessian data requirements. This trend confirms that scarce second-order derivative information refines local PES curvature while maintaining the overall accuracy of the energy and force.

Our Hessian-informed training protocol achieves substantial absolute performance gains. The original AlphaNet model trained on the full HORM dataset reported a Hessian MAE of 0.415 \unit{eV/\angstrom^2} and an eigenvalue MAE of 3.887 \unit{eV/\angstrom^2}~\cite{HORM}. In contrast, by pre-training on the semi-empirical HORM-xTB-Hess dataset and fine-tuning on 100,000 configurations with our framework, the errors are reduced to 0.105 and 0.360 \unit{eV/\angstrom^2}, respectively. 
To explore full potential of the framework, we carry out a more extensive training by pre-training on T1x-xTB-Hess ($\sim$9.7M) followed by fine-tuning on the full HORM datasets, which enables the model push the Hessian and eigenvalue MAE down to 0.075 and 0.20 \unit{eV/\angstrom^2}.

To further identify the contributions of each component in the training protocol, we performed an ablation study comparing effect of performing pre-training and different loss functions~(Fig.~\ref{fig:mol_react}c). Our results demonstrate that the Hutchinson loss significantly outperforms standard column-wise evaluation~(labeled as ``one-col.'' in Fig.~\ref{fig:mol_react}c), yielding a substantial reduction in both Hessian and eigenvalue errors. Furthermore, the integration of Hessian pre-training consistently improves model performance, with the combination of Hutchinson loss and xTB-level pre-training achieving the lowest overall errors. These findings demonstrate that coupling proper stochastic projection loss with Hessian pre-training is essential for capturing high-fidelity curvature with scarce labels.

Moving beyond simple error metrics, we assessed the utility of Hi-MLIP on practical downstream applications.

Transition state search and Gibbs free energy calculation are essential tasks in first-principle investigation of chemical reactions. For transition state search, we follow the ReactBench~\cite{ReactBench} benchmark, containing 960 reactions, to evaluate the ability of Hi-MLIP in locating transition state~(Details in Sec.~\ref{sec:TS_search}). Furthermore, the Gibbs free energies of the reactant/product and the corresponding transition state~(TS) are calculated using the rigid-rotor harmonic-oscillator approximation to benchmark with the DFT results. 

We construct Hi-MLIP for transition state relevant tasks by pre-training on T1x-xTB-Hess, and then perform energy-based sampling for high-fidelity DFT configurations. To better capture the curvature around the TS, we found that high-energy configurations, which predominantly represent TS regions, play a more vital role than low-energy configurations more close to reactants and products. As shown in Fig.~\ref{fig:mol_react}d, the model trained on high-energy subsets significantly outperforms those trained on low-energy subsets. Notably, the inclusion of Hessian information~(E/F/H) enhances the success rate of locating TS compared to models trained solely on energy and forces~(E/F), especially in regimes where data is scarce. 
Moreover, we find that even a minimal amount of Hessian data can significantly improve the performance of the model.
As shown in Fig.~\ref{fig:mol_react}e, AlphaNet trained on a Top 5\% high-energy subset maintains a success rate comparable to the full-data baseline even when the Hessian sampling is reduced to a mere 0.1\% of the 5\% subset, demonstrating the extreme data efficiency of our protocol.

Finally, we evaluated the performance of the model in thermodynamic property prediction by calculating MAE for Gibbs free energies of reactants, products~(R/P), and transition states~(TS) on a randomly selected 124 reactions(Fig.~\ref{fig:mol_react}f, and see Methods for details). Here we present the result predicted by the baseline models from original HORM work~\cite{HORM}. After including Hessian labels, the MAE of free energy for equilibrium structures~(R/P) is improved closely to chemical accuracy~(defined as 0.043 eV or 1 kcal/mol). However, the MAE for TS structures remains far beyond acceptable accuracy for correct predictions of reaction barriers.
On the other hand, our most data-efficient model, utilizing 5\% DFT EF labels of full dataset and 0.1\% Hessian labels of the subset, closely approaches the threshold of chemical accuracy for both R/P and TS structures. 
These results demonstrate our protocol can accurately reproduce thermodynamic properties at DFT accuracy while reducing requirement of high-fidelity Hessian labeling to $0.0005\%$ of the full set.

Beyond the data-efficiency, we also explore the possibility of further improving the accuracy. We 
pre-train Hi-MLIP on the enlarged T1x-xTB-Hess~($\sim$9.7M) and fine-tune it on full HORM DFT datasets. 
The result is depicted in Fig.~\ref{fig:mol_react}f labeled as "best", which reaches the MAE of $0.01$$\text{eV}$, demonstrating the systematical improvability of our approach for the accuracy demand.


\subsection*{Phonon Anharmonicity in Hydrides}

\begin{figure*}[ht]
 \centering
 \includegraphics[width=0.75\textwidth]{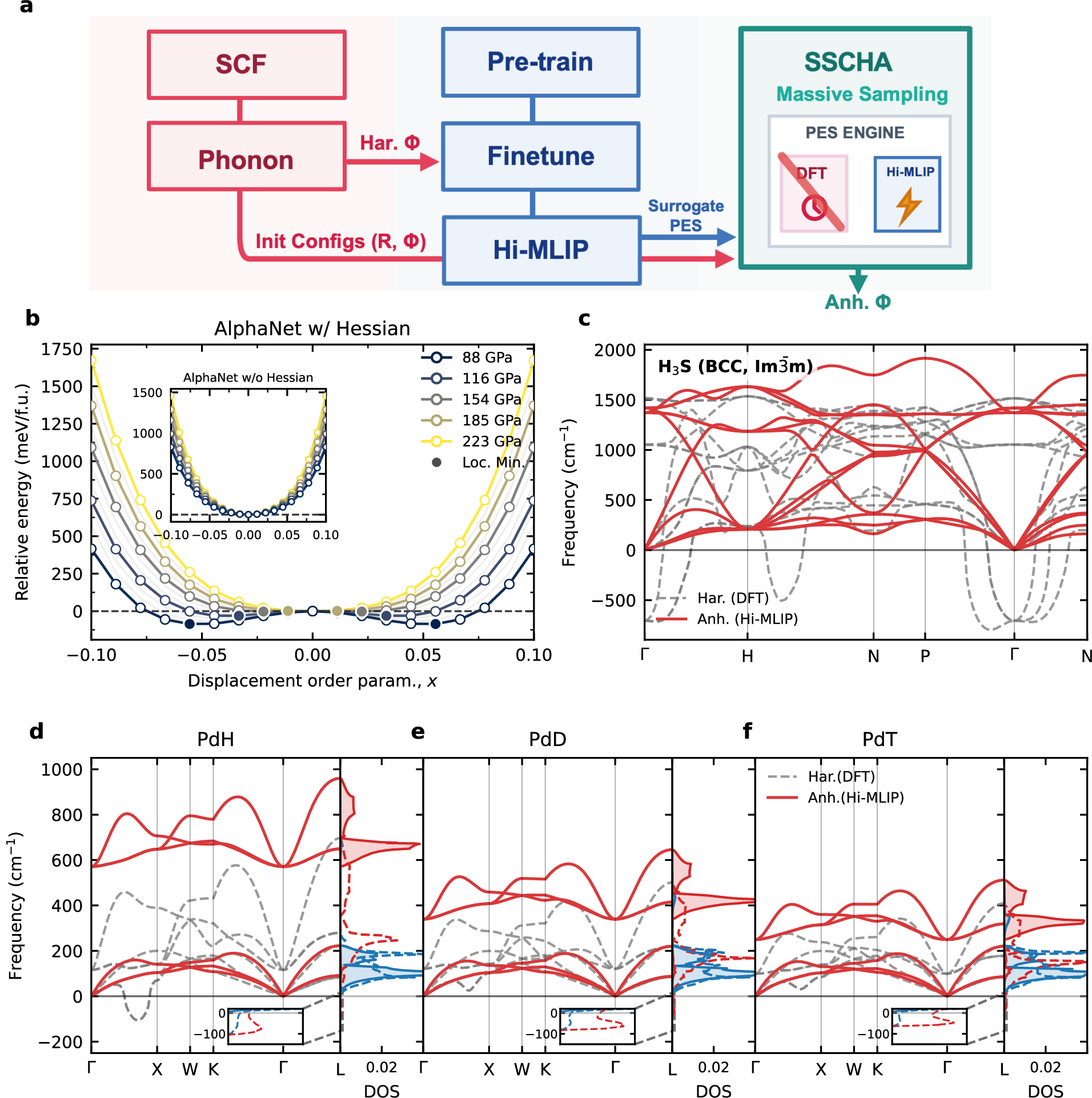}
\caption{\textbf{Anharmonic effects on the dynamic stability of hydrosulfide and palladium hydrides.}
        \textbf{(a)}  Workflow for predicting anharmonic dynamic matrix~($\Phi$) using the SSCHA framework enhanced by Hessian-informed training protocol~(HINT). Red boxes represent DFT calculations, blue boxes represent HINT part, and green box represents SSCHA part. \textbf{(b)} Energy variation of the H$_3$S structure as a function of the displacement order parameter (where H atoms move slightly toward S atoms, indicating a second-order phase transition) under different pressures. The calculations are performed using the AlphaNet model with Hessian fine-tuning~(without Hessian fine-tuning for inset). Solid dark dots indicate the local minima.
       \textbf{(c)} Harmonic and anharmonic phonon dispersions of the $Im\bar{3}m$ phase of H$_3$S at 150 GPa. The gray dashed lines represent the harmonic results calculated by DFT, while the red solid lines represent the anharmonic results optimized by SSCHA using Hi-MLIP.
        \textbf{(d)},\textbf{(e)},\textbf{(f)}
        Harmonic and anharmonic phonon dispersions with the projected phonon density of states~(DOS) for PdH, PdD, and PdT. For DOS, dashed and solid lines represent harmonic and anharmonic calculations, respectively. The blue and red filled regions denote the contributions from the Pd atoms and the H/D/T isotopes.
}
    
\label{fig:anharmonic}
\end{figure*}

While the harmonic approximation is usually sufficient for isolated molecules near equilibrium geometries as we studied in the preceding sections, anharmonicity becomes non-negligible in condensed phases characterized by strong dynamical fluctuations, such as systems featuring light elements, high temperatures, or structural phase transitions~\cite{wangFull2026}. In these systems, anharmonicity plays the pivotal role in ensuring the dynamic stability and governing macroscopic properties observed in experiments, such as ultralow thermal conductivity in thermoelectrics~\cite{zhaoUltralow2014}, phase stabilization in ferroelectrics~\cite{zhongPhase1994, tadanoFirstPrinciples2018}, and high-temperature superconductivity for hydrides under pressure~\cite{h3s, erreaQuantum2020}. However, incorporating these effects via first-principles anharmonic treatments is often computationally intractable, limiting their use in large-scale and high-throughput calculations.

The Stochastic Self-Consistent Harmonic Approximation~(SSCHA)~\cite{SSCHA2014, SSCHA2021} is a rigorous framework for capturing anharmonic dynamics.
However, performing SSCHA optimization purely with DFT 
presents a heavy computational burden, which often demands tens of thousands of costly supercell force evaluations. 
To overcome this computational bottleneck, 
we combine the HINT protocol and the resulting Hi-MLIP with the conventional SSCHA method. 
As illustrated in Fig.~\ref{fig:anharmonic}a, the workflow initiates with standard DFT calculations to obtain the SCF ground state and the subsequent harmonic phonon dispersions within a supercell. Within the HINT framework, we employ an AlphaNet model~\cite{alphanet} pre-trained on the large-scale, lower-fidelity OMAT24 dataset~\cite{OMat24}. For the given hydride system, this model is fine-tuned using high-fidelity data characterized by a larger basis set and tighter convergence thresholds. This fine-tuning set comprises the exact Hessian matrix from the initial harmonic step, augmented by 30 high-precision single-point evaluations of the unit cell using a dense k-mesh, a process representing only a fraction of the computational cost of supercell evaluations. By reusing the initial Hessian required by the SSCHA workflow, HINT requires zero additional cost for this Hessian, enabling the model to serve as an efficient and accurate surrogate for subsequent SSCHA optimizations.

The structural transition in \ce{H3S} under pressure emphasizes the role of Hessian supervision in accurately characterizing PES curvature~\cite{h3s}.
According to the DFT calculation under harmonic approximation, a low-symmetry $R3m$ phase is predicted to be the ground state. Yet combined synchrotron X-ray diffraction and Raman spectroscopy confirm the high-symmetry $Im\bar{3}m$ phase near 150 GPa. 
The presence of the high-symmetry phase is because the strong zero-point energy of hydrogen becomes non-negligible and stabilize the $Im\bar{3}m$ structure~\cite{goncharovStable2017}.

This stabilization involves a second-order phase transition where H atoms displace toward S atoms, characterized by a double-well potential energy surface. As illustrated in Fig.~\ref{fig:anharmonic}b, an AlphaNet model fine-tuned \textit{without} Hessian information incorrectly predicts a simple quadratic single well potential energy surface. While this quadratic landscape might coincidentally stabilize the high-symmetry phase through error cancellation, it completely misses the correct physical picture. In contrast, our Hessian-fine-tuned model accurately reproduces the double-well landscape, confirming that the stabilization is captured through correct physics rather than numerical artifacts. 
Moreover, the resulting the anharmonic phonon dispersions for the $Im\bar{3}m$ phase at 150 GPa (Fig.~\ref{fig:anharmonic}c), eliminating the severe imaginary frequencies seen in the harmonic results, achieving good agreement with the experimental observation.

The HINT-SSCHA framework similarly eliminates the dynamic instabilities of palladium hydride~(\ce{PdH}) and its isotopes~(\ce{PdD}, \ce{PdT}) within harmonic approximation. As shown in Fig.~\ref{fig:anharmonic}d, the harmonic DFT calculations yield imaginary frequencies and heavily underestimate optical phonon modes. The anharmonicity introduced via SSCHA optimization eliminates these instabilities, hardens the hydrogen-related optical branches and exhibits a narrowing trend of phonon band gap from PdH to PdT.

\begin{figure*}[ht]
 \centering
 \includegraphics[width=0.75\textwidth]{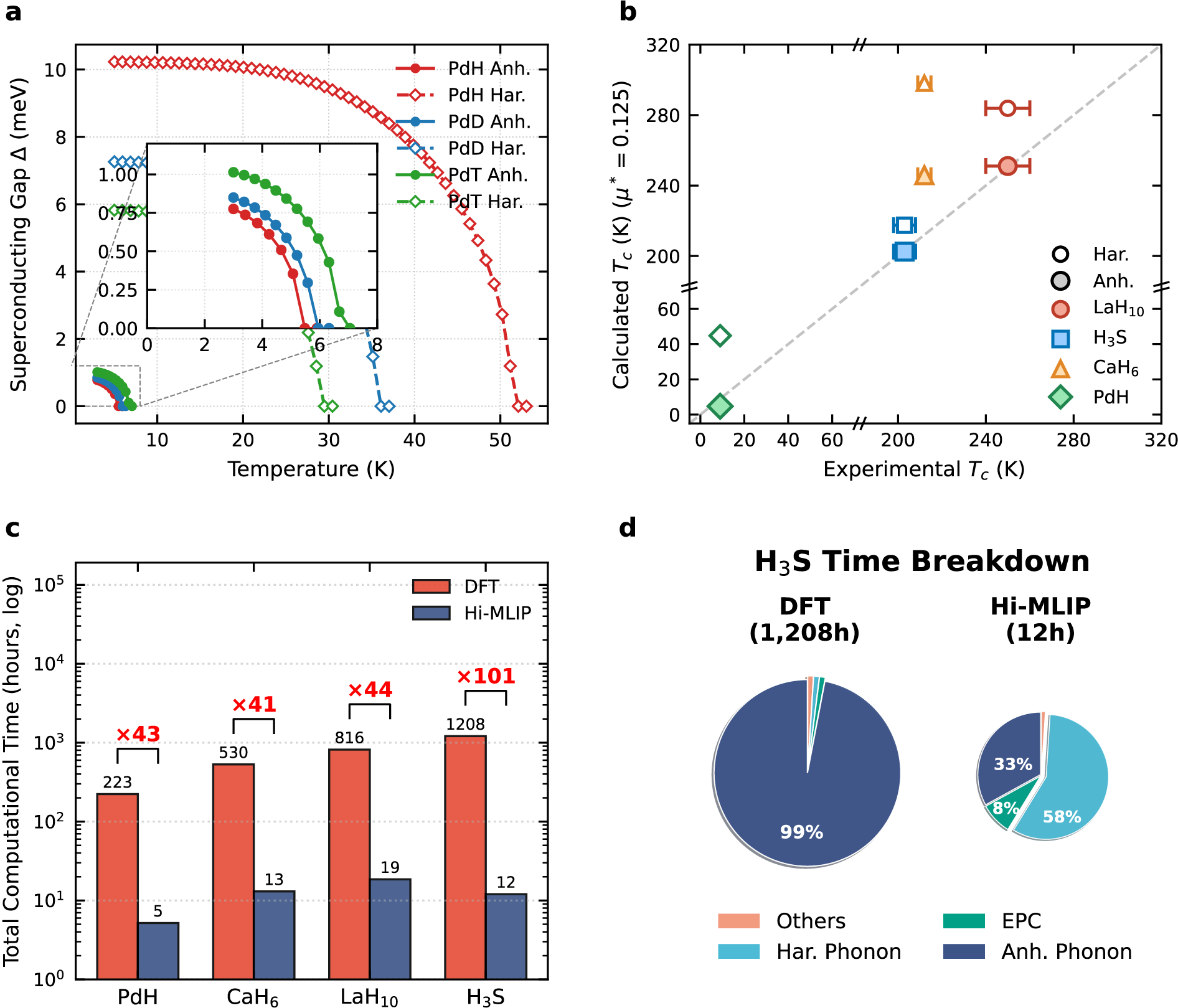}
\caption{\textbf{Efficient evaluation of phonon anharmonicity and superconducting critical temperatures via Hi-MLIP.}
        \textbf{(a)}  Temperature dependence of the superconducting gap ($\Delta$) for PdH, PdD, and PdT, calculated by solving the Migdal-Eliashberg (ME) equations. Open diamonds with dashed lines indicate harmonic results, while solid circles with solid lines denote anharmonic results. The inset provides a magnified view of the low-temperature region for PdH, PdD and PdT. \textbf{(b)} Superconducting critical temperature ($T_c$) for LaH$_{10}$ at 200 GPa~(red), H$_3$S at 150 GPa~(blue), CaH$_6$ at 180 GPa~(brown) and PdH at 1 atm~(green). Open and solid symbols correspond to anharmonic and harmonic results, respectively. The dashed gray line indicates perfect agreement between theory and experiments~\cite{lah10,h3s,cah6,pdh}. \textbf{(c)} Comparison of computational efficiency between the Hi-MLIP approach and the pure DFT approach~(estimated) for anharmonic calculations across four hydrides. 
        \textbf{(d)} 
        The pie charts illustrate the percentage of time consumed by each computational step for \ce{H3S}.}
    
\label{fig:superconduct}
\end{figure*}

Beyond ensuring dynamical stability, anharmonicity critically modulates the macroscopic properties of functional materials. 
In pressurized hydrides, these effects significantly renormalize the phonon frequencies and, consequently, the superconductivity in these systems. To quantify this impact, we evaluate the superconducting critical temperature~($T_c$) using the Allen-Dynes modification of the McMillan formula~(see Sec.~\ref{sec:superconduct} for details).

Specifically, for \ce{PdH} and its isotopes (\ce{PdD}, \ce{PdT}), incorporating anharmonicity not only resolves the dynamical instabilities but also accurately captures the inverse isotope effect in superconductivity (Fig. \ref{fig:superconduct}a).
Due to the lighter mass of hydrogen~(\ce{H}), its zero-point fluctuations sample a broader region of the anharmonic potential well compared to deuterium~(\ce{D}) and tritium~(\ce{T}), which paradoxically renormalizes the frequencies. By solving the Migdal-Eliashberg equations with these anharmonic inputs, we reproduce the hierarchy of the superconducting gap ($\Delta$) and $T_c$ for the Pd-H/D/T systems, observed experimentally by low-temperature resistance measurements and electron tunneling spectroscopy~\cite{schirberConcentration1974a,eichlerTunneling1975,biancoEnhanced2026}.

Beyond capturing the correct physical trends in the \ce{PdH} series, this approach provides quantitative predictions for $T_c$ in various high-pressure hydrides.
As shown in Fig.~\ref{fig:superconduct}b, the $T_c$ calculated for \ce{LaH10}~\cite{lah10}, \ce{H3S}~\cite{h3s}, \ce{CaH6}~\cite{cah6} and \ce{PdH}~\cite{pdh} using the McMillan Allen-Dynes equation align closely with experimental measurements. By explicitly incorporating phonon anharmonicity, the predicted $T_c$ all show improvements over harmonic ones. Notably, at a commonly adopted Coulomb pseudopotential of $\mu^* = 0.125$, the discrepancy between our anharmonic predictions and experimental measurements is less than 5 K for most systems, with the exception of \ce{CaH6}.

To quantify this efficiency gain, we compare HINT-SSCHA against the conventional DFT-SSCHA cost. While standard DFT-based procedures typically demand $5,000$ to $60,000$ supercell evaluations for the tested hydrides, HINT effectively removes this bottleneck by substituting them with Hi-MLIP surrogates Fig.~\ref{fig:superconduct}d. The finetuning itself is performed on the Hessian matrix from phonon step~(Fig.~\ref{fig:superconduct}a), augmented by 30 configurations by cheap unit-cell single points calculations.
Thus, the SSCHA process does not grow significantly with system complexity. Consequently, for the four hydrides investigated, the entire SSCHA workflow adds only $2.5$ to $4$ hours to the initial harmonic calculation. We note that while previous studies have also utilized MLIPs for SSCHA~\cite{mlip-sscha}, our development no longer heavily relied on active learning to sample phase space and construct potentials from scratch, which typically still requires an initial first-principles phonon calculation followed by approximately 500 costly supercell calculations.

\section{Discussion}

As demonstrated, Hi-MLIP delivers more accurate predictions for complex thermodynamic behavior, transition-state kinetics, and highly anharmonic systems.
This improvement results from its enhanced capability to capture local curvature of PES, made possible by moving beyond conventional training schemes that rely exclusively on energy and force~(E/F) data.
In order to reliably construct Hi-MLIPs, HINT protocol addresses computational bottlenecks through Hessian pre-training, targeted configuration sampling, curriculum learning and stochastic Hessian loss formulation.
As a result, the requirement for high-fidelity DFT Hessian labels is reduced by up to four orders of magnitude, substantially lowering the barrier for realistic applications.
A concurrent work by Rodriguez et al.~\cite{rodriguezProjected2026} also investigated stochastic Hessian loss formulation and demonstrated its advantage over full Hessian loss. 

Looking forward, the evolution of Hi-MLIP will benefit from faster and more universally accurate methods to generate pre-training datasets. Furthermore, beyond simple fine-tuning, integrating HINT with advanced multi-fidelity training methods~\cite{kimDataEfficientMultifidelityTraining2025,batatiaCrossLearningElectronic2025} promises to further enhance representation learning. Broadly, the extreme data efficiency established here is not restricted to DFT accuracy. Most importantly, this extreme data efficiency dictates that DFT need not be the ceiling. It opens a potential pathway to train MLIPs using highly scarce labels from computationally demanding wavefunction theory, such as coupled cluster theory~\cite{riplingerEfficient2013,yePeriodicLNOCC2024,huangImprovable2025} or quantum Monte Carlo~\cite{foulkes2001quantum,li2024computational,scherbela2025accurate}. Finally, by enabling the practical incorporation of accurate PES curvature into machine learning potential, this unlocks the ability to routinely evaluate vibrational properties across catalysts, drug crystals and functional materials, thus bridging atomistic theory and experimental observations. Furthermore, extending this framework to accurately capture higher-order derivatives of the PES represents a vital next step for the explicit treatment of strong anharmonic effects. Such advancements are essential for the high-fidelity prediction of complex macroscopic observables governed by phonon-phonon interactions, including lattice thermal conductivity, thermal expansion, and temperature-dependent anharmonic spectra~\cite{hiphive, phanh, spectra}.

\section{Methods}\label{sec:methods}

\subsection*{Hessian-informed Training Protocol}
\label{sec:hutinson}
The Hessian-informed training~(HINT) protocol is composed of four major parts, namely the Hessian pre-training and configuration sampling to achieve the data-efficiency; curriculum learning and stochastic Hessian for the training efficiency. 

\textbf{Machine Learning Potential Architecture.}
HINT is general for all kinds of MLIPs. In this work, we employ AlphaNet~\cite{alphanet}, a state-of-the-art equivariant graph neural network, as the backbone for learning the PES. 
Unlike conventional equivariant models that rely on computationally expensive tensor products of irreducible representations based on spherical harmonics, AlphaNet adopts a local-frame-based approach. 
This architecture achieves $
\text{SO(3)}$ symmetry by constructing complete equivariant local frames and scalarizing geometric information within these frames. 
Such a design allows for high-precision while maintaining the computational efficiency.

\textbf{Hessian Pre-training.}
To mitigate the formidable computational cost of high-fidelity DFT Hessians, we employ a pre-training strategy using multi-fidelity data. This approach establishes a foundational PES curvature representation using low-cost data before further fine-tuning.
For the molecular systems, we construct two extensive auxiliary low-fidelity datasets, T1x-xTB-Hess~($\sim$9.6M) and HORM-xTB-Hess~($\sim$1.7M). These datasets inherit the molecular configurations from the original Transition1x~\cite{T1x} and HORM~\cite{HORM} datasets, but all labels, including energy, forces, and the full $3N \times 3N$ Hessian matrices, are re-evaluated at the GFN2-xTB level~\cite{gfn2-xtb}. 
For material systems, we substitute xTB level data with OMat24~\cite{OMat24} due to the inherent limitations of tight-binding methods in describing metallic elements.

\textbf{Configuration Sampling.}
After the low-fidelity representation of PES is established, we enter a data sampling stage to identify the most informative configurations for the subsequent high-fidelity fine-tuning. Several different sampling strategies are employed in different scenarios.

\textit{Weighted-Local Density}.
To ensure the model effectively captures rare but physically informative configurations, we employ the weighted-local density~(WLD) sampling. In contrast to uniform sampling which tends to over-sample structures in regions with dense data, WLD ensures a more balanced sampling by up-weighting under-represented configurations in sparse regions.

For each configuration $i$ in the latent space~(e.g. SOAP descriptors~\cite{SOAP}), we estimate its local density $\rho_i$ using the average Euclidean distance to its $k$-nearest neighbors~($k$-NN). To enhance structural diversity, the sampling probability $P_i$ is defined to be inversely proportional to the local density, and thus directly proportional to the average neighbor distance:\begin{equation}P_i \propto \frac{1}{k} \sum_{j \in \text{NN}(i)}^k \| \mathbf{z}_i - \mathbf{z}_j \|^2,\end{equation}where $\mathbf{z}_i$ denotes the embedding vector. This weighting scheme ensures that high-density regions are less likely to be sampled, while geometrically unique structures in sparse regions receive higher sampling weights.

\textit{Energy-based Sampling}.
For transition state searching tasks, we performed energy-based sampling on the HORM dataset~\cite{HORM} to evaluate the importance of Hessian information across different PES regions. Configurations were categorized into high-energy~(close to transition regions) and low-energy~(close stable basins) subsets based on their relative energies within stoichiometric isomers. Within each subset, we maintained full E/F labels while restricted Hessian labels to a randomly sampled fraction~(e.g. 0.1\% of the given subset).

\textit{Random Sampling for Anharmonicity}.
Since the SSCHA workflow requires an initial supercell Hessian, we reuse this data for fine-tuning at zero additional computational cost. We augment this dataset with 30 unit-cell configurations, generated by applying random displacement vectors to the atomic positions based on a zero-mean Gaussian distribution with an amplitude~(standard deviation) of 0.4\unit{\angstrom}. By performing these auxiliary calculations in the unit cell rather than the supercell, we obtained high-fidelity training labels at an affordable computational cost. All auxiliary labels were computed using a converged $k$-point grid.

\textbf{Curriculum Learning.}
To mitigate training instability and overfitting caused by the extreme scarcity of Hessian labels~(e.g., 0.1\%), we implement a dynamic growing weight strategy during high-fidelity fine-tuning. Without dynamic weighting, Hessian loss during training tend to exhibit a ``U-shaped'' trajectory that initially decreases but then increases quickly, as the model overfits limited Hessian data. To address this, we treat Hessian-informed learning as a curriculum, formulated within the total loss:\begin{equation}\mathcal{L}_{total}(t) = w_E \mathcal{L}_E + w_F \mathcal{L}_F + \lambda(t) \mathcal{L}_{H},\end{equation}According to our tests using AlphaNet, we set $w_E = 4$ and $w_F = 100$. The Hessian weight $\lambda(t)$ is linearly increased from $w_0 = 0$ at $t_{start}$ to $w_H = 0.1$ at $t_{end}$:\begin{equation}\lambda(t) = w_0 + (w_H - w_0) \cdot \frac{t - t_{start}}{t_{end} - t_{start}}.\end{equation}This strategy ensures the model first focuses on tuning the global PES inherited from low-fidelity pre-training before gradually introducing high-fidelity Hessian information. Consequently, we prevent overfitting by limited Hessian labels during fine-tuning.

\textbf{Stochastic Projected Hessian Loss.}
To mitigate the $\mathcal{O}(N^2)$ memory and computational overhead of the full Hessian matrix $\mathbf{H} \in \mathbb{R}^{3N \times 3N}$, we define the training objective as a stochastic projection loss. Instead of explicitly reconstructing the full Hessian matrix in the loss function $\mathcal{L}_{\text{H}} = \| \mathbf{\tilde{H}} - \mathbf{H} \|_F^2$, we optimize Hessian-vector products. The loss is calculated directly as the mean squared error~(MSE) between the predicted and reference projected vectors:
\begin{equation}
    \mathcal{L}_{\text{H}} = \frac{1}{3N} \| \mathbf{\tilde{H}}\mathbf{z} - \mathbf{H}\mathbf{z} \|_2^2
\end{equation}
where $\mathbf{z} \in \mathbb{R}^{3N}$ is a stochastic projection vector.
The conventional approach employs coordinate basis vectors $\{\mathbf{e}_i\}_{i=1}^{3N}$, where $(\mathbf{e}_i)_j = \delta_{ij}$. This scheme is equivalent to randomly select the individual columns of the Hessian matrix, as utilized in previous works~\cite{HORM,PFTphonon}. The alternatively way is to construct $\mathbf{z}$ from a Rademacher distribution~($\pm 1$ with equal probability), satisfying the conditions of mean zero ($\mathbb{E}[\mathbf{z}] = \mathbf{0}$) and unit covariance ($\mathbb{E}[\mathbf{z} \mathbf{z}^\top] = \mathbf{I}$) as inspired by Hutchinson trace estimation~\cite{hutchinsonStochastic1989}. In our tests, these Rademacher projections exhibit better training accuracy compared to conventional coordinate basis vectors.
In our implementation, we use a sample size of $m=5$ random vectors per batch to achieve a robust estimation of the loss.

\subsection*{Transition State Search and Thermodynamic Analysis}\label{sec:TS_search}
The localization and characterization of transition states~(TS) are central to modeling chemical reaction kinetics and mechanisms. In this work, we employ the workflow in ReactBench~\cite{ReactBench} for TS search and validation at full MLIP level. There are in total 960 reactions in ReactBench. The workflow begins with the geometric optimization of the input reactant and product structures. Subsequently, an initial TS guess is generated using growing string method~(GSM)~\cite{zimmerman2015single}. The node with highest energy among the reaction path is further optimized using restricted-step rational-function-optimization~(RS-I-RFO)~\cite{besalu1998automatic}. Finally, the intrinsic reaction coordinate~(IRC) calculation is performed to verify in both reaction directions, to ensure that the forward and reverse IRC endpoints are exact the input reactant and product. If IRC verified is passed, then the TS is classified as ``success'' in Fig.~\ref{fig:mol_react}d,e.

The thermodynamic analysis shown in Fig.~\ref{fig:mol_react}f is performed through PySCF package~\cite{PySCF_1,PySCF_2}. 
To ensure a consistent comparison and eliminate the influence of failed TS optimized by MLIP, the geometric structure optimized in DFT is used when calculating the Gibbs free energy.

\subsection*{Anharmonic Phonon Calculation}
\textbf{Stochastic Self-Consistent Harmonic Approximation.}
To rigorously account for the quantum zero-point motion of the lattice and anharmonic thermal fluctuations, the stochastic self-consistent harmonic approximation (SSCHA) method is employed. The SSCHA method is based on the variational principle, approximating the true system by introducing a parameterized trial harmonic Hamiltonian $\mathcal{H}_0(\boldsymbol{\mathcal{R}}, \boldsymbol{\Phi})$. Its core objective is to minimize the variational free energy functional of the system:
\begin{equation}
\mathcal{F}_{SSCHA}[\boldsymbol{\mathcal{R}}, \boldsymbol{\Phi}] = \text{tr}[\rho_{\mathcal{H}_0}\mathcal{K}] + \langle V(\mathbf{R}) \rangle_{\rho_{\mathcal{H}_0}} - T\mathcal{S}[\rho_{\mathcal{H}_0}]
\end{equation}
where the variational parameters $\boldsymbol{\mathcal{R}}$ represent the average atomic positions~(centroid structure), and $\boldsymbol{\Phi}$ is the trial effective force constant matrix (or dynamical matrix). $\mathcal{K}$ is the kinetic energy operator, $T$ is the temperature~(set to $0.0$ K in this study to capture purely quantum zero-point effects), $\mathcal{S}$ is the entropy of the system, $V(\mathbf{R})$ is the true Born-Oppenheimer PES, and $\langle \cdot \rangle_{\rho_{\mathcal{H}_0}}$ denotes the ensemble average over the trial harmonic density matrix $\rho_{\mathcal{H}_0}$. The initial $\boldsymbol{\Phi}$ matrix is obtained from standard harmonic phonon calculations and subsequently symmetrized and forced to be positive definite to eliminate imaginary frequencies, ensuring the validity of the initial probability distribution.

Hi-MLIP is integrated as the energy and force evaluator for SSCHA. In each sampling iteration, an ensemble of 500 random supercell configurations satisfying the quantum harmonic Gaussian distribution is generated based on the current $\boldsymbol{\mathcal{R}}$ and $\boldsymbol{\Phi}$. Then Hi-MLIP engine efficiently computes the total energies, atomic forces, and stresses for all configurations in the ensemble. These statistical results are directly utilized to analytically evaluate the free energy gradients with respect to $\boldsymbol{\mathcal{R}}$ and $\boldsymbol{\Phi}$. The maximum number of population sampling iterations is restricted to 30.

The optimization proceeds via simultaneous updates of both atomic positions~($\alpha_{\boldsymbol{\mathcal{R}}}$) and dynamical matrix~($\alpha_{\boldsymbol{\Phi}}$) using a step size of 0.002 to ensure stability in the gradient descent process. 
The computational efficiency is further enhanced by the Kong-Liu reweighting technique~\cite{kong1994sequential,liu2009large}, which allows for expectation value updates without immediate resampling. 
New configurations are generated only when the effective sample ratio drops below 0.5, indicating a significant deviation from the original trial distribution. The convergence is reached when the free energy gradients fall below 0.1\% of the statistical noise threshold, as determined by a stochastic meaningful factor of 0.001. The entire self-consistent relaxation process alternates between ensemble sampling and functional minimization until both the centroid coordinates and the effective dynamical matrix are fully converged within the statistical limits.

\textbf{Superconducting Temperature Prediction.}\label{sec:superconduct}
The superconducting critical temperature~($T_c$) is determined using the Allen-Dynes modification of the McMillan formula:
\begin{equation}
    T_c = \frac{f_1 f_2 \omega_{\ln}}{1.2} \exp \left[ -\frac{1.04(1 + \lambda)}{\lambda - \mu^*(1 + 0.62\lambda)} \right],
\end{equation}
where $\lambda$ represents the dimensionless electron-phonon coupling (EPC) constant, $\omega_{\ln}$ is the logarithmic average frequency derived from the harmonic or anharmonic phonon spectrum, and $\mu^*$ is the effective Coulomb repulsion parameter. The terms $f_1$ and $f_2$ represent the strong-coupling and spectrum shape correction factors, respectively, which depend on $\lambda$, $\mu^*$, $\omega_{\ln}$, and the root-mean-square phonon frequency $\bar{\omega}_2$. 

The core parameters $\lambda$ and $\omega_{\ln}$ are directly computed from the Eliashberg spectral function, $\alpha^2F(\omega)$, through the following integrals:
\begin{equation}
    \lambda = 2 \int_{0}^{\infty} \frac{\alpha^2F(\omega)}{\omega} \, d\omega,
\end{equation}
\begin{equation}
    \omega_{\ln} = \exp \left[ \frac{2}{\lambda} \int_{0}^{\infty} \frac{\alpha^2F(\omega)}{\omega} \ln(\omega) \, d\omega \right].
\end{equation}
The anharmonic phonon frequencies are calculated using the SSCHA method accelerated by Hi-MLIP. The Hi-MLIP employs AlphaNet as the backend, which is pre-trained on OMat24~\cite{OMat24}. The fine-tuned stage is performed by reusing the harmonic Hessian generated by SSCHA for initialization without additional computational cost. This data is augmented with 30 unit-cell configurations perturbed by random Gaussian displacements zero-mean and 0.4\unit{\angstrom} amplitude (standard deviation).
\subsection*{Dataset Composition}

\textbf{T1x dataset}: The Transition1x~(T1x) dataset is a large-scale repository specifically designed for molecular reactions~\cite{T1x}. T1x contains 9.6 million configurations derived from 10,013 organic reactions involving C, H, N, and O atoms. Unlike traditional datasets sampled from equilibrium MD trajectories, T1x structures are sampled along Nudged Elastic Band (NEB)~\cite{sheppard2008optimization} pathways at the $\omega$B97x/6-31G(d)~\cite{chai2008systematic,ditchfield1971self} level of theory, capturing critical regions of the PES such as transition states and reaction intermediates.

\textbf{HORM dataset}: 
HORM is a comprehensive subset of the Transition1x~(T1x) dataset~\cite{T1x} and RGD1 dataset~\cite{RGD1} with additional curvature information~\cite{HORM}. HORM contains 1,725,362 molecular configurations, each labeled with energy, forces, and the full Hessian matrix at $\omega$B97X/6-31G(d)~\cite{chai2008systematic,ditchfield1971self} level of theory. HORM is used for fine-tuning in HINT for molecular systems studied in this work. The validation set of HOMR is used to calculate MAE in Fig.~\ref{fig:mol_react}a-c.  
In this work, HORM serves as the data source for fine-tuning stage in HINT for molecular systems, and its validation set is used to evaluate the MAE reported in Fig.~\ref{fig:mol_react}a-c.

\textbf{T1x-xTB-Hess and HORM-xTB-Hess datasets}: We introduce two datasets using the same configurations from T1x~($\sim$ 9.6M)~\cite{T1x} and HORM~($\sim$ 1.7M)~\cite{HORM} at GFN2-xTB level with full Hessian matrix for each configuration. These two datasets are used for pre-training stage in HINT for molecular systems studied in this work. These two datasets will be released publicly with this work. 

\textbf{OMat24 dataset}: The Open Materials 2024~(OMat24) dataset is a massive-scale repository for inorganic crystalline materials~\cite{OMat24}. It comprises approximately 118 million configurations spanning a vast chemical space across the periodic table, with energies, forces, and stress tensors calculated at the PBE~\cite{PBE} level of theory with periodic boundary conditions and the projector augmented wave~(PAW) pseudopotentials~\cite{PAW} implemented in VASP~\cite{VASP}. The Hubbard U corrections for oxide and fluoride materials containing Co, Cr, Fe, Mn, Mo, Ni, V, or W are used in constructing dataset~\cite{OMat24}. The OMat24 dataset serves as the pre-train data in HINT for hydrides studied in this work.

\textbf{Hydride Phonon Data Samples.} To accurately characterize the hydrogen-dominated lattice dynamics, DFT calculations are conducted via Quantum ESPRESSO~\cite{QE1,QE2} at the PBE level~\cite{PBE}. The electronic iterations are converged to within $10^{-12}$ Ry~($1.36 \times 10^{-11}$ eV), and a plane-wave basis set with a cutoff of $70-90$ Ry~($950 - 1225$ eV) is used. For reciprocal space integration, we enforce a $k$-point spacing equivalent to a real-space periodicity of over $100 \text{ \AA}$ in each direction. This dense sampling is essential for capturing the fine details of the Fermi surface and the electron-phonon coupling matrix elements. These phonon data samples are using in fine-tuning stage for hydrides studied in this work.

\section*{Acknowledgments}
The authors acknowledge Dr. Hang Li and ByteDance Seed AI for Science teams for their invaluable support. H.X. is supported by the National Natural Science Foundation of China (Grant No. 22525302). J.C. is supported by the National Key R$\&$D Program of China under Grant No. 2021YFA1400500.

\bibliography{refs}

\end{document}